\documentclass[conference]{IEEEtran}
\IEEEoverridecommandlockouts
\usepackage{cite}
\usepackage{graphicx}
\usepackage{textcomp}
\usepackage{booktabs}
\usepackage{multirow}
\usepackage{array}
\usepackage{multicol}
\usepackage{url}
\usepackage[utf8]{inputenc}
\usepackage{caption}
\usepackage{amsmath, amssymb}

\usepackage{algorithm}
\usepackage{algpseudocode}

\usepackage{xcolor}
\usepackage{tikz}
\usepackage{subfig}

\usepackage[framemethod=TikZ]{mdframed}
\usepackage{tcolorbox}
\tcbuselibrary{breakable,skins}

\usepackage{enumitem}

\usepackage{fontawesome}

\usepackage{listings}
\definecolor{annotationGray}{RGB}{128,128,128}
\definecolor{boxBackground}{RGB}{245,245,245}

\newtcolorbox{annotationbox}[1][]{
    enhanced,
    before skip=10pt,
    after skip=10pt,
    colback=boxBackground,
    colframe=annotationGray,
    arc=3mm,
    boxrule=0.5pt,
    left=10pt,
    right=10pt,
    top=8pt,
    bottom=8pt,
    fontupper=\normalsize,
    title style={
        left color=annotationGray,
        right color=annotationGray,
        font=\bfseries\sffamily\color{white}
    },
    overlay={
        \node[anchor=west, text=white, font=\bfseries\sffamily]
        at ([xshift=10pt]frame.north west) {Annotation};
    }
}
\def\BibTeX{{\rm B\kern-.05em{\sc i\kern-.025em b}\kern-.08em
    T\kern-.1667em\lower.7ex\hbox{E}\kern-.125emX}}
\begin{document}

\title{Floorplan2Guide: LLM-Guided Floorplan Parsing for BLV Indoor Navigation}

\author{
    \IEEEauthorblockN{Aydin Ayanzadeh and Tim Oates}
    \IEEEauthorblockA{
        \textit{University of Maryland, Baltimore County}\\
        Baltimore, Maryland, USA\\
        \{aydina1, oates\}@umbc.edu
    }
}
  \maketitle

\begin{abstract}

Indoor navigation remains a critical challenge for people with visual impairments. The current solutions mainly rely on infrastructure-based systems, which limit their ability to navigate safely in dynamic environments. We propose a novel navigation approach that utilizes a foundation model to transform floor plans into navigable knowledge graphs and generate human-readable navigation instructions. Floorplan2Guide integrates a large language model (LLM) to extract spatial information from architectural layouts, reducing the manual preprocessing required by earlier floorplan parsing methods. Experimental results indicate that few-shot learning improves navigation accuracy in comparison to zero-shot learning on simulated and real-world evaluations. Claude 3.7 Sonnet achieves the highest accuracy among the evaluated models, with 92.31\%, 76.92\%, and 61.54\% on the short, medium, and long routes, respectively, under 5-shot prompting of the MP-1 floor plan. The success rate of graph-based spatial structure is 15.4\% higher than that of direct visual reasoning among all models, which confirms that graphical representation and in-context learning enhance navigation performance and make our solution more precise for indoor navigation of Blind and Low Vision (BLV) users.  

\end{abstract}


\section{Introduction}
According to a recent report, approximately 2.2 billion individuals worldwide live with visual impairment~\cite{WHO}. Individuals with visual impairments encounter significant barriers to independent mobility, including challenges with travel, orientation, and acquiring spatial information, necessitating safe and reliable navigation support. In addition to traditional mobility aids such as white canes and guide dogs, recent technological advancements utilizing artificial intelligence provide individuals with visual impairments enhanced mobility and independence~\cite{srinivasaiah2024turn, thandu2024chatbot}.

Recent progress in Multimodal Large Language Models (MLLMs) has improved navigation reasoning performance by integrating visual data with textual information, enabling high-level path planning~\cite{zhou2023navgpt, xu2024flame}. This research builds upon outdoor navigation systems, where MLLMs can interpret and describe visual information to assist Blind and Low-Vision (BLV) users in outdoor environments, demonstrating accurate performance in open-world scenarios. Outdoor navigation systems function reliably, as they rely on GPS to determine users' positions.

\begin{figure}[t]
\centering
\includegraphics[width=\columnwidth , height=0.37\textheight]{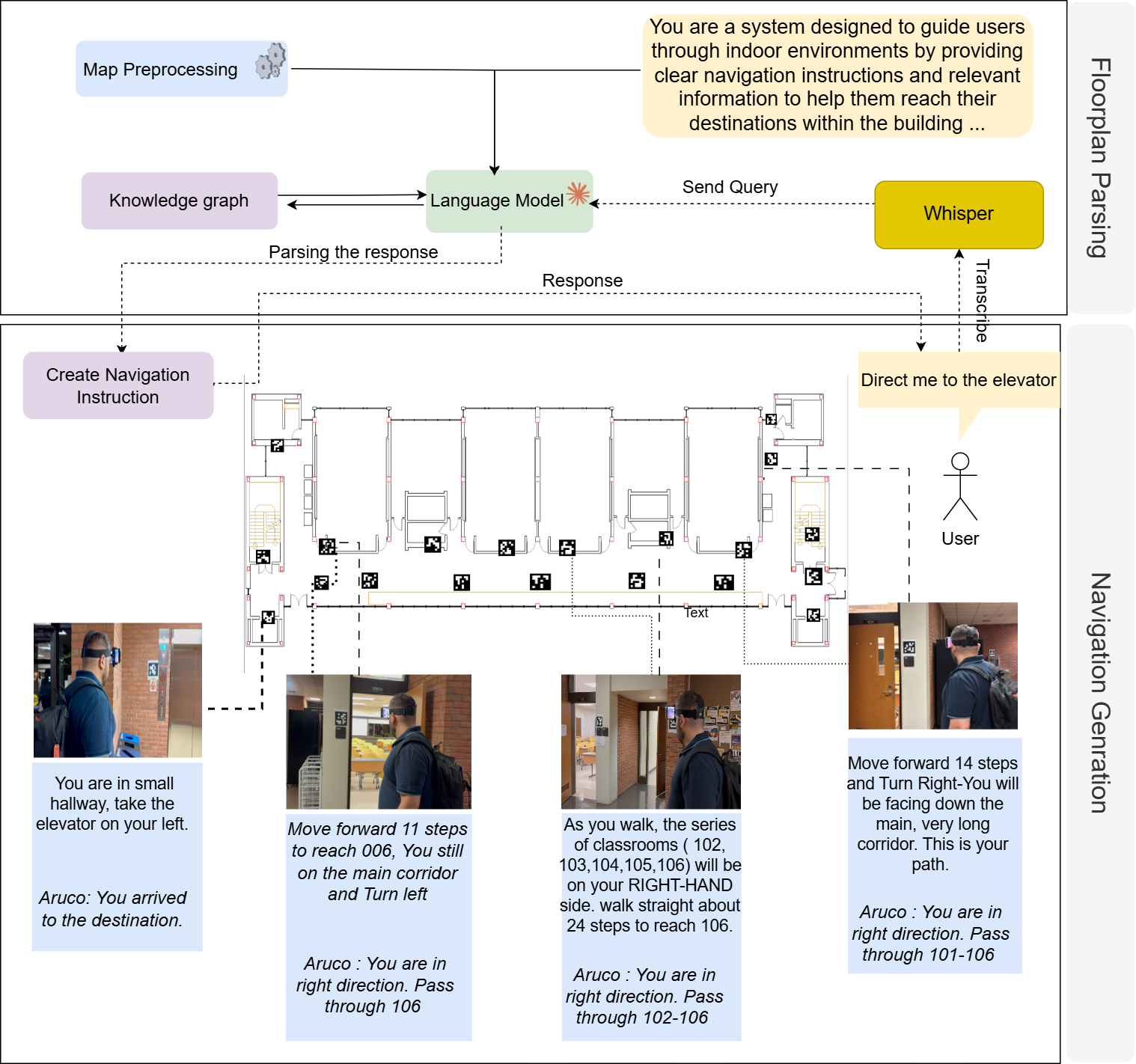}
\caption{Architecture of the LLM-based indoor navigation system. The framework parses floorplans into a spatial knowledge graph interpreted by the language model, which generates navigation instructions from user queries (e.g., "Direct me to the elevator").}
\label{fig:workflow_paper}
\end{figure}

\begin{figure*}[t]
    \centering
    \includegraphics[width=\textwidth, height=0.40\textheight]{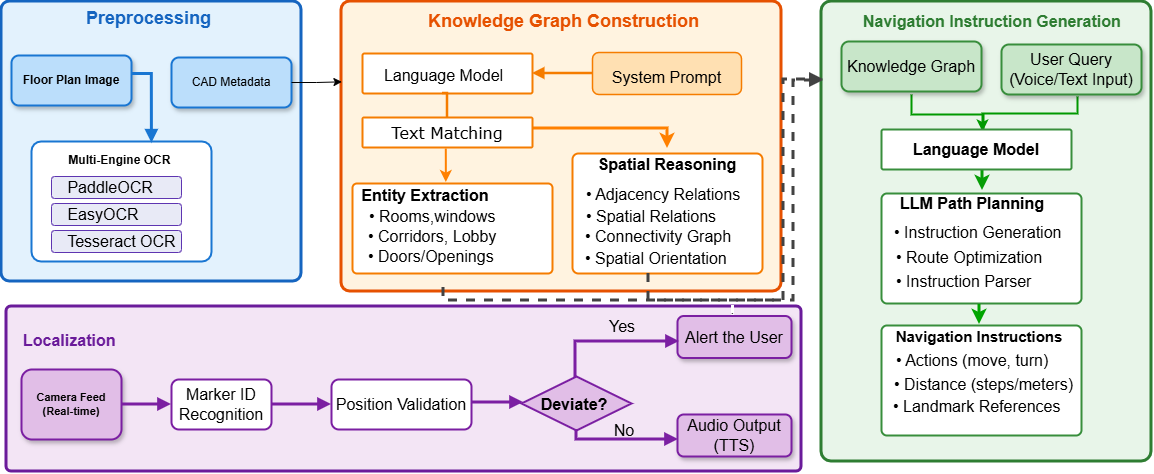}
    \caption{Overall workflow of the proposed Floorplan2Guide framework. The system takes a floorplan image and a user query (e.g., ``I want to use the restroom'') as input. The preprocessing module extracts textual and geometric features using OCR and visual analysis, while the LLM constructs and validates a knowledge graph enriched with ArUco marker information. Finally, the navigation module generates step-by-step, context-aware instructions. During localization, the system alerts the user if they deviate from the route; otherwise, it repeats the last navigation instruction.}
    \label{fig:workflow}
\end{figure*}

Floorplan analysis is a critical phase in computer vision for indoor navigation. Traditional methods utilize image processing and graph-based algorithms to extract the main components of floorplans, including walls and doors~\cite{floor1, floor2, floor3, deep_floor1}.  While graph embeddings excel on link prediction tasks~\cite{m1,m2}, node classification, graph visualization, etc. Integrating language models with knowledge graph systems can enhance knowledge representation and improve the reasoning capabilities of LLMs\cite{KGLLM}. Traditional methods rely on geometric and structural heuristics, which often require extensive preprocessing and manual correction. Recent advances have enabled new deep learning methods and multimodal reasoning for floorplan analysis, where LLMs can interpret floorplans to identify entities and relationships~\cite{map1, Navip} directly. Indoor localization serves as the foundation for indoor navigation systems. Infrastructure-based methods, such as BLE beacons~\cite{In1, In2, In3, In4, In5} and RFID tags~\cite{RFID1,RFID2}, offer precise positioning but require costly installation and maintenance. In contrast, lightweight localization approaches employ sensors and cameras with minimal infrastructure requirements, such as printed fiducial markers~\cite{sens1,sens2}, and rely on computer vision and machine learning for localization~\cite{ml1,ml2}. However, these models depend on large, building-specific datasets and lack generalization to new environments~\cite{Navip}. Fiducial marker-based methods offer efficient localization with easier deployment and better scalability. However, purely vision-based deep learning navigation systems require extensive retraining for each environment, are computationally intensive, and often fail to adapt dynamically to unseen layouts or lighting variations, limiting their real-world applicability. As shown in Fig.~\ref{fig:workflow_paper}, this study proposes a novel framework that adapts floorplan understanding for indoor navigation by converting floorplan imagery into spatial knowledge graphs, which enable LLMs to generate context-aware navigation instructions for visually impaired individuals~\cite{abidi2024review, map1}.

Our contributions can be summarized as follows:

\begin{enumerate}

\item We introduce a novel solution that utilizes foundation models to parse floorplans by transforming the spatial information into a knowledge graph that enables accurate spatial parsing to extract semantic information and generate structured representations for navigation tasks.

\item We employ few-shot learning in our approach to improve navigation performance as evaluated using the standard Success Rate (SR). Few-shot learning leads to more consistent spatial reasoning and improves navigation reliability without additional fine-tuning.

\item We validate generated navigation instructions in a real-world environment at the Department of Math \& Psychology Building (MP-1) at UMBC, and demonstrate that the majority of navigation routes generated by our approach are successful. Instructions are more effective when they specify the number of steps and the distance to be traversed, making them more accessible to BLV users.

\end{enumerate}

The rest of the paper is organized as follows: Section~\ref{sec:related} provides an overview of the state of the art in AI-driven assistive systems for individuals with BLV. Section~\ref{sec:method} provides an in-depth discussion of the methodology adopted for the proposed system, examining its crucial components and implementation approach. Section~\ref{sec:results} presents the experimental results and findings on the system's performance, along with its benefits. Section~\ref{sec:limitations} discusses the limitations in this research that can be addressed in future work. Finally, conclusions are drawn in Section~\ref{sec:conclusion} based on the findings presented in this work.

\section{Related Work}
\label{sec:related} 

\subsection{Floorplan Understanding}

\begin{figure}[t]
    \centering
    \includegraphics[width=0.98\columnwidth]{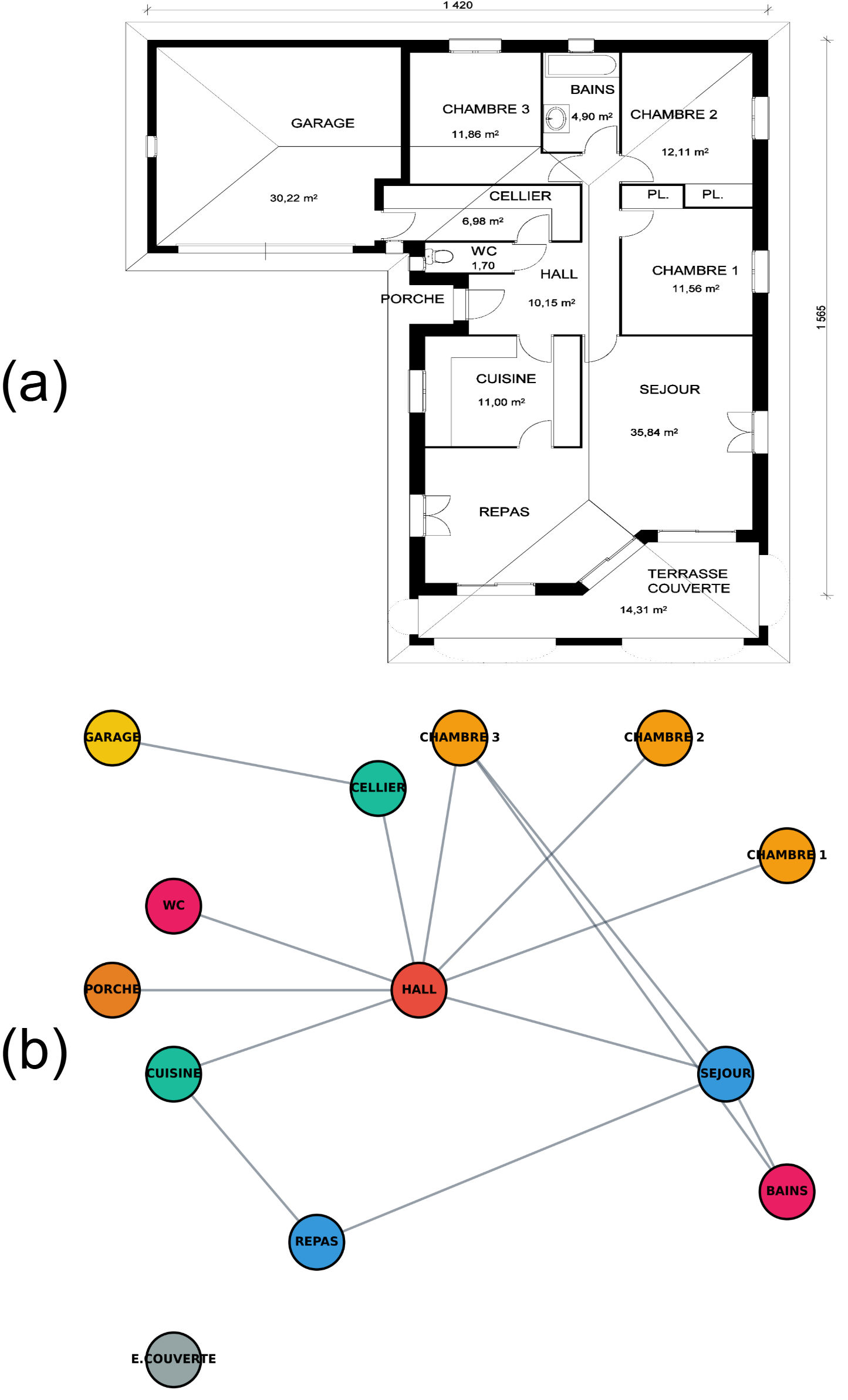}
    \caption{(a) Example floor plans from the CVC-FP dataset and (b) corresponding adjacency graph representations.}
    \label{fig:spatial_graph}
\end{figure}

Several studies have explored ways to create navigation routes from building floorplans, representing them as graphs with edges and nodes \cite{floor1, floor2, floor3}. Earlier work focused on analyzing floorplans found on information boards in shopping malls or building entrances to extract walkable areas. We build on this work by developing a method that analyzes floorplans for navigation and provides turn-by-turn instructions for blind individuals using foundation models. Our system extracts key information such as intersections, destinations, and their connections, then generates a node map to build a spatial knowledge graph for navigation tasks. Fig.~\ref{fig:spatial_graph} shows how each floorplan from the CVC-FP\cite{CVC} dataset gets converted into an adjacency graph representing room connectivity and walkable areas.

The conversion process examines every room pair in a floorplan, creating an edge when their polygons are close enough or a door connects them. Edges can represent doors or passages, while nodes store attributes like centroids, geometry, and room types. Deep learning has opened new possibilities for automatically creating graph representations from architectural floorplans. Chen and Stouffs \cite{deep_floor1} proposed a framework using ensemble learning and semantic segmentation to generate attributed adjacency graphs from floorplan images. Renz et al. \cite{renz2022graph} developed methods for converting floorplan images into precise mathematical graph structures for floorplan analysis. A recent study shows that vision-language models can parse floor plans directly, without traditional preprocessing \cite{VLN_parse}. While their main focus is on robots and VLN tasks, we take this further by evaluating floorplan parsing on CVC-FP and additional datasets, applying LLMs for both semantic analysis and node map creation. Our approach extends floorplan parsing to LLM-guided understanding, enabling context-aware navigation. Coffrini et al. \cite{map1} proposed using a multimodal LLM to interpret floorplans as input images with visual prompting, though their work still needs real-world validation. As demonstrated in our Experiment III, we demonstrate that graph-based reasoning improved navigation accuracy by up to 15.4\% compared to direct visual reasoning on zero-shot prompting.

\subsection{Multimodal Language Models}

Assistive technologies for people with BLV have made significant progress in recent years, particularly with AI-powered solutions \cite{Int7, Int8, Int9}. These technologies combine computer vision and natural language processing to provide visual interpretation. Commercial applications, such as Microsoft's Seeing AI\cite{seeingAI}, have gained widespread adoption by using cameras for object identification. More recently, BeMyEyes\cite{bemyeye} integrated OpenAI GPT-4\cite{gpt-4} for image-to-text tasks, helping blind users navigate independently in daily activities. Smartphone-based object detection systems now support blind users in independent travel, with applications ranging from vending machines to finding accessible restrooms. The NaVIP \cite{Navip} system takes a different approach, requiring extensive labeled image data with exact 6DoF pose information to work within specific building areas. Commercial applications, such as NaviLens~\cite{navilens} and Zapvision~\cite{zapvision}, along with hybrid positioning systems, demonstrate that fiducial marker–based methods are effective in public settings and improve accessibility for BLV individuals. Inspired by fiducial marker-based approaches, we employ ArUco-based fiducial markers and integrate them into our proposed navigation approach to address the limitations of image-centric approaches~\cite{Navip}\cite{ml1} which need to be retrained across different environments, which makes our solution more robust in dynamic indoor settings.

\begin{figure*}[ht]
\centering
\includegraphics[width=\textwidth, height=0.40\textheight]{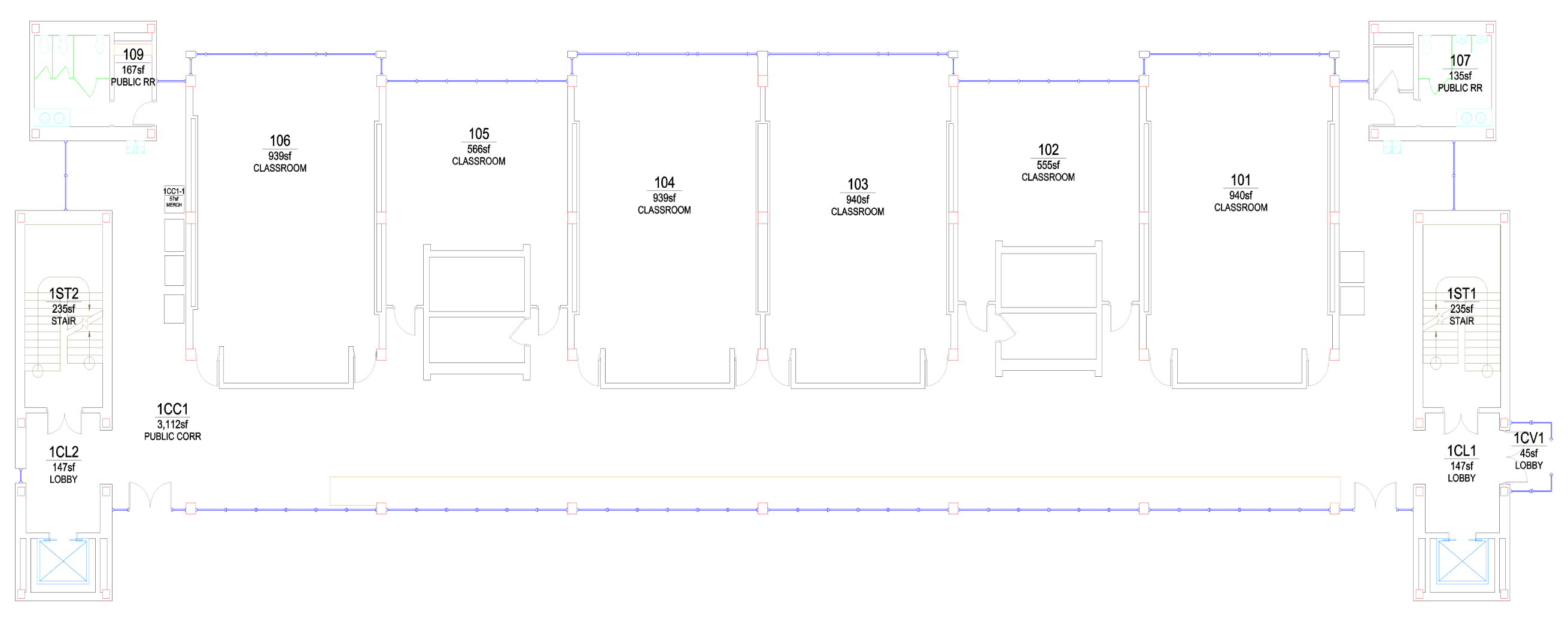}
\caption{The MP-1 indoor floor plan used in our experiments to evaluate the navigation task, including room layouts, corridors, and key structural elements that were analyzed by the proposed method.}
\label{fig:mp1_floorplan_1}
\end{figure*}

\section{Methodology}
\label{sec:method}
 We discuss four key components of our approach: floorplan understanding, semantic parsing and graph representation, navigation instruction, and localization. 
 
\subsection{Problem Formulation}
\label{sec:problem}

Indoor navigation for users with visual impairments requires robust semantic and spatial understanding of the built environment, often derived from visual inputs, such as floorplan images. The challenge addressed in this work is the automatic construction of a spatial knowledge graph from a given architectural floorplan image, enabling reliable pathfinding, spatial reasoning, and downstream navigation assistance. Given a floorplan image $I$ and a navigation task $S$, our goal is to extract a spatial knowledge graph $G = (V, E, A, M_\eta, W, S)$ where $V$ is the set of nodes, each corresponding to a distinct room or place, and $E$ represents the edges of the graph that encode navigability, adjacency, or connections between rooms or areas. The adjacency matrix $A$ encodes binary connectivity between nodes. The term $M_\eta$ is architectural metadata for nodes and edges, such as the kind of door and its spatial relation and other floorplan elements. Moreover, by extracting spatial information from the floorplan, our pipeline records the centroids, relative positions, and spatial relations between nodes and floorplan elements. This formulation depicts the demonstrated graph, which  not only demonstrates topological connectivity but also the semantic and spatial context necessary for indoor navigation and visual reasoning. We employed fiduacial marker based approach which unlike the

\subsection{System Overview }

To parse the floorplan, our system uses LLMs to transform static floorplans into machine-readable navigation data through a structured multi-stage analysis. The overall process consists of consecutive stages that collectively convert a visual floor map into graph-based, navigable representations.

\subsubsection{\textbf{Preprocessing}}
High-quality floorplan images are first processed to capture both visual and textual information. The floorplan extraction stage includes Optical Character Recognition (OCR) to extract textual data such as room numbers, labels, and annotations. Although language models can interpret text from images, OCR achieves higher accuracy in recognizing small or distorted characters. The extracted text is then supplied to the language model, enabling it to enrich its floorplan understanding and produce more reliable semantic labels.

\subsubsection{\textbf{System Prompt}} The system prompt is hard-coded for the language model and is applied uniformly to all users. The prompts for generating the knowledge graph and for navigation instructions are demonstrated in the appendix. It contains comprehensive instructions for extracting room info, guidelines for properly extracting the node map per the graph rules, and explanations of the accessibility between two nodes. Moreover, we provide guidelines to eliminate errors in the extraction of semantic information from the floorplan. We use several rounds of refinement to prevent significant errors and improve the model's overall performance in transforming the floorplan into a navigable graph.

\begin{figure*}[t]
    \centering
    \includegraphics[width=\textwidth]{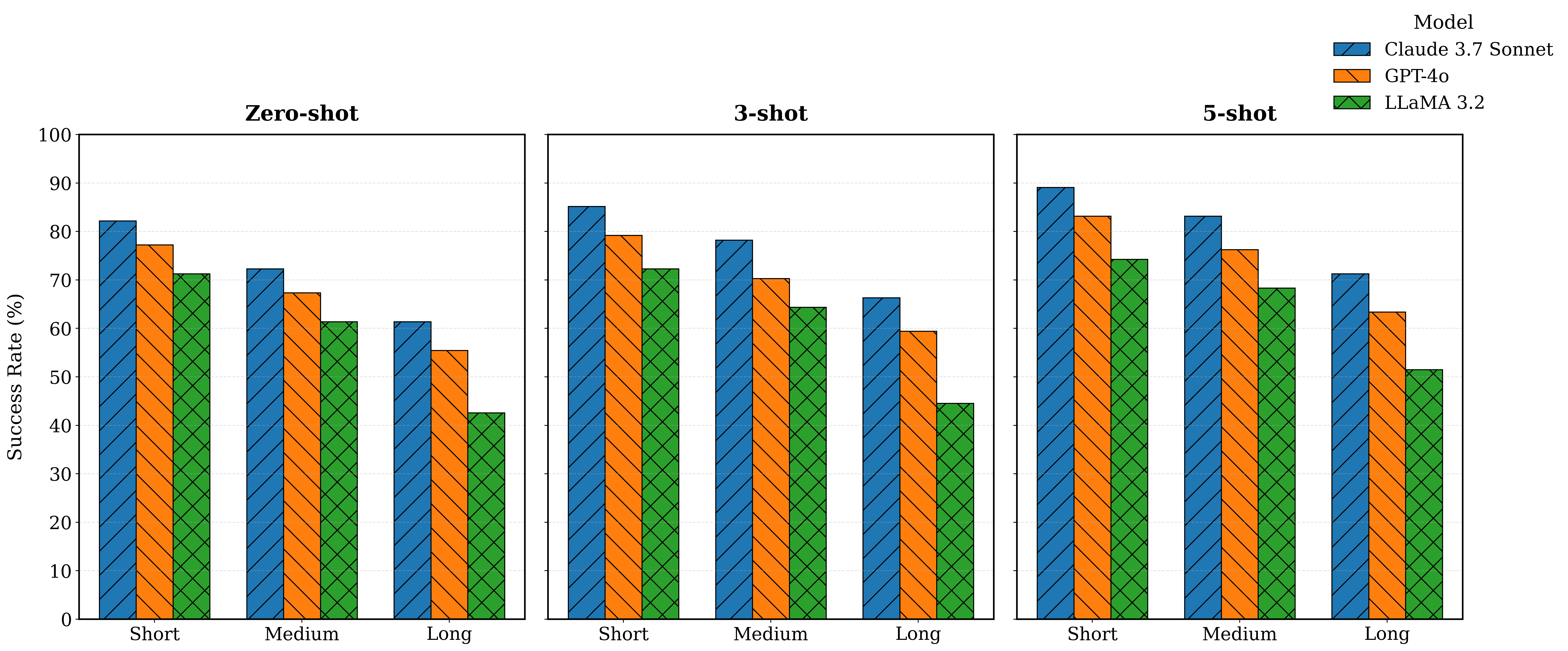}
    \caption{Performance of three LLM models, including GPT-4o, Claude 3.7 Sonnet, and LLaMA 3.2 Vision-Instruct  11B, on zero-shot and few-shot prompting on the CVC-FP dataset. A total of 303 navigation instructions were evaluated as test cases. The same prompt was used for all models, which indicates the success rate on navigation per route type.}
    \label{fig:CVC_final_plot}
\end{figure*}

\subsubsection{\textbf{Spatial Knowledge Graph Generation}}

Through multimodal reasoning, the model integrates visual parsing and textual context to comprehend the overall building layout. Each detected element is assigned a unique node ID containing metadata on its geometry, functionality, and accessibility. The visual map remains synchronized with a digital spatial knowledge graph that serves as the foundation for navigation and path generation. Algorithm~\ref{alg:skg_extraction} represents the procedure of knowledge graph extraction from the floorplan. The extracted entities are transformed into a graph structure $G = (V, E, A, M_\eta, W, S)$, where nodes $V$ represent spatial entities (e.g., rooms and intersections), edges $E$ denote walkable paths, and weights $W$ capture distance and accessibility information. Based on the representation, an adjacency matrix is created to encode spatial relationships to create a node map for the floorplan. Each node stores both geometric and semantic attributes derived from the model’s multimodal analysis. A scale factor converts floorplan measurements into real-world dimensions, improving distance estimation and navigation accuracy. The system labels spatial regions and models typical movement patterns, allowing it to generate natural, human-like navigation instructions from the starting point to a specified destination.

\begin{algorithm}[ht]
\caption{Spatial Knowledge Graph Extraction}
\label{alg:skg_extraction}
\begin{algorithmic}[1]
\Require floorplan image $I$, door locations $D$, MLLM $\mathcal{M}$
\Ensure Spatial knowledge graph $G = (V, E, A, M_\eta, W, S)$

\State $B \gets \emptyset$
\For{$o \in \{\text{PaddleOCR}, \text{EasyOCR}, \text{Tesseract}\}$}
    \State $B \gets B \cup \text{FilterByConfidence}(\text{OCR}_o(I), \tau)$
    \If{$|B| \geq N_{\min}$} \textbf{break} \EndIf
\EndFor

\State $P \gets \text{BuildPrompt}(I, B, D)$
\State $(V, A, M_\eta) \gets \text{ParseMLLMOutput}(\mathcal{M}(P))$

\ForAll{$v \in V$}
    \State $v.\text{box} \gets$ \Call{MatchSpatialBox}{$v, B, D$}
    \State $S[v] \gets \text{ComputeGeometry}(v.\text{box})$
\EndFor

\State $E \gets \{(v_i, v_j) : A[i][j] = 1, i < j\}$
\State \Return $G = (V, E, A, M_\eta, W, S)$
\end{algorithmic}
\end{algorithm}

\subsubsection{\textbf{Navigation Instruction}}

The graph generated in the prior step serves as the main input to the language model for the navigation task. Unlike direct visual reasoning, where the floorplan image is processed directly, our approach feeds the structured knowledge graph to the language model along with the system prompt. This graph-based input provides explicit spatial relationships and room connectivity, enabling more accurate and consistent generation of navigation instructions. Finally, the language model provides a sequence of steps for high-level planning, generating instructions that provide optimized directions based on the floorplan's knowledge graph, which stores the distance to each node. In this study, we employed a language model to generate optimized directions rather than using shortest-path algorithms. This approach demonstrates the potential of language models to generate step-by-step instructions by parsing each step and providing human-readable directions to BLV users.

\subsubsection{\textbf{ArUco Marker-based localization}}

The localization process requires that ArUco markers \cite{aruco} be linked to the graph's node IDs. Each marker has a unique ID that points to the generated graph, which links what the camera sees to the information in our system. We place these markers at different parts of the building, including entrances, intersections in the main corridor, and rooms. The user needs to head to the ArUco and navigate to the nearest ArUco marker for the localization, and the system uses its map to determine their location. To validate the user's current position, the user needs to scan an ArUco marker. The system then localizes the user using the nearest marker and uses that to determine their exact location. If the marker aligns with a node in the navigation instruction, the system informs the user by either repeating the last direction or by confirming that you are on the right track. This feedback loop helps the user navigate to the destination more confidently.  The ArUco marker also checks whether the user has deviated from the pathway. If the user deviates from the route, the system alerts the user.  In this study, we did not use ArUco markers for the rerouting task; instead, they were used only as checkpoints to estimate the user’s position. The system enables  BLV users to be more confident through its combination of ArUco with knowledge graphs.

\subsection{Implementation Details}

The system uses the phone's camera to capture video input and the microphone to receive audio, making interaction easier. The system transmits both voice recordings and captured images to a server for processing, which generates comprehensive navigation instructions. We utilize Whisper~\cite{whisper} to transcribe user queries, which the language model then processes. We employ GPT-4o, Claude 3.7 Sonnet, and the LLaMA 3.2 Vision-Instruct   11B~\cite{llama3.2} model for knowledge graph creation and navigation reasoning. All experiments on the MP-1 floorplans, which are indicated in Fig~\ref{fig:mp1_floorplan_1}, were performed on a MacBook Pro equipped with an Apple M1 Pro processor with a 10-core CPU, 16-core GPU, and 16 GB RAM. A head-mounted camera streamed live video input to the local system. This phone camera is operated for the ArUco detection module to scan and recognize markers deployed in the environment, each linked to a specific node ID in the knowledge graph to assist navigation to the destination. We employed Success Rate (SR), which is a standard metric to evaluate navigation performance, defined as  $ SR = N _ {success}/N _ {total} $ where $N_{success}$ indicates the number of successful trials and $N_{total}$ indicates the total number of trials for navigation.

\begin{figure*}[t]
    \centering
    \includegraphics[width=\textwidth]{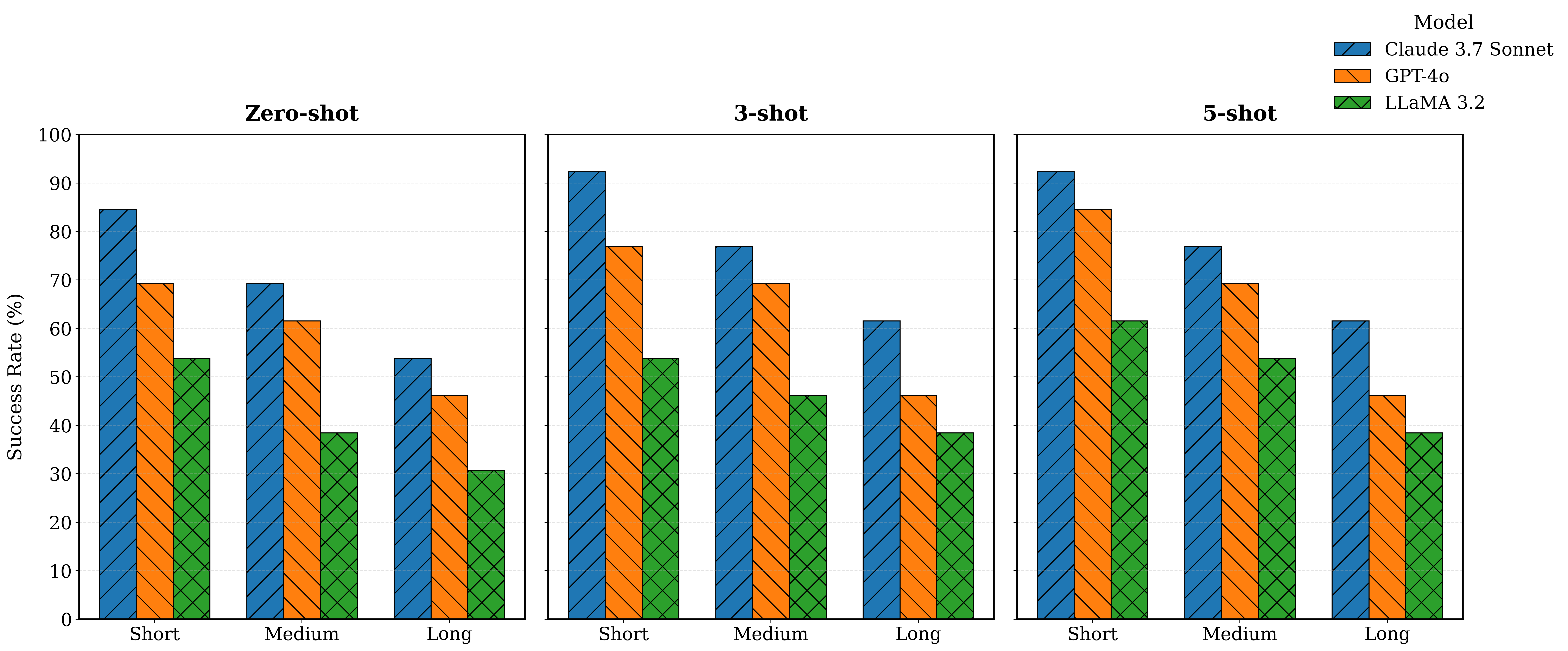}
    \caption{Performance of GPT-4o, Claude 3.7 Sonnet, and LLaMA 3.2 Vision-Instruct on zero-shot, 3-shot, and 5-shot prompting across the UMBC Math \& Psychology Floor 1. Each bar indicates the success rate (\%) over three navigation types, including short, medium, and long.}
    \label{fig:mp1_plot}
\end{figure*}

\section{Experiments}
\label{sec:results}

\subsection{Experiment I}

The first experiment was conducted on the CVC-FP dataset~\cite{CVC} to evaluate our system’s baseline performance using standardized floorplan images. The CVC-FP is a publicly available benchmark for structural floorplans analysis that includes diverse architectural layouts with annotated rooms, doors, and walls. As stated in Fig.~\ref{fig:spatial_graph}, a representative sample from the CVC-FP dataset is shown with their corresponding adjacency graph representations. Each floorplan included three navigation tasks of varying lengths (short, medium, and long). As shown in Fig.~\ref{fig:CVC_final_plot}, all models improved with few-shot prompting. Claude~3.7~Sonnet achieved the highest success rates, reaching 84.6\%, 76.9\%, and 61.5\% for short, medium, and long paths, respectively. The observed gain over zero-shot prompting ranged from 6\% to 9\% depending on path complexity. These results confirm that providing few-shot prompting as a contextual cue enables more accurate route inference and fewer directional errors, particularly in complex multi-turn routes.

\subsection{Experiment II}

We conducted real-world evaluations on the UMBC Math \& Psychology Building to examine the practical performance of our navigation framework. As in the first experiment, we employed Claude 3.7 Sonnet, GPT-4o, and LLaMA 3.2 Vision-Instruct under zero-shot, 3-shot, and 5-shot prompting conditions. A trial was considered successful when the generated instructions accurately guided the user from the starting point to the destination without colliding with doors, walls, or other elements. Obstacle avoidance was excluded since dynamic detection is not part of this study. As illustrated in Fig.~\ref{fig:mp1_plot}, all models exhibit improvement from zero-shot to few-shot prompting, with the best results achieved under 5-shot conditions. GPT-4o has improved from 69.23\% to 84.62\%  on short routes, and conversely, its SR does not increase when applying few-shot learning, neither with 3-shot nor with 5-shot prompting on long routes. Although in some cases we do not have improvement from 3-shot to 5-shot learning, which shows the lower improvement. However, in most cases, few-shot learning improves SR across different route categories, as these results confirm that few-shot prompting enhances spatial reasoning coherence and stability in real-world indoor navigation scenarios. Among the employed methods, Claude 3.7 Sonnet achieves the highest SR for all path types, with 61.54\%, 76.92\%, and 92.31\% on long, medium, and short route types, respectively.

\subsection{Experiment III}

In the third experiment, we evaluated the impact of explicit graph representation on navigation performance. We compared our graph-based approach with a direct vision-language reasoning baseline that uses the same visual input but without intermediate graph construction. We evaluate graph representations for zero-shot learning to isolate the method's impact, without adding any auxiliary techniques, in both versions. As presented in Table~\ref{tab:graph_comparison}, graph-based spatial structuring significantly improved performance across all models. Claude~3.7~Sonnet achieved 84.62\%, GPT-4o 76.9\%, and LLaMA~3.2~Vision-Instruct 61.5\% under the graph-based configuration in the MP-1 floorplan. These findings demonstrate that explicit topological encoding through knowledge graphs improves route planning, reduces hallucination errors, and strengthens spatial consistency in LLM-based indoor navigation.

\begin{table}[htbp]
\centering
\caption{Impact of Graph Construction across different Route Types on MP-1 in Zero-shot prompting. Numbers in parentheses indicate successful 
trials out of 13 total routes.}

\label{tab:graph_comparison} 
\footnotesize
\setlength{\tabcolsep}{5pt}
\renewcommand{\arraystretch}{1.10}
\begin{tabular}{|l|c|c|c|}
\hline
\textbf{Model (Condition)} & \textbf{Short (\%)} & \textbf{Medium (\%)} & \textbf{Long (\%)} \\
\hline
Claude 3.7 (w/o Graph) & 76.92 (10) & 61.54 (8) & 38.46 (5) \\
Claude 3.7 (w/ Graph)  & \textbf{84.62 (11)} & \textbf{69.23 (9)} & \textbf{53.85 (7)} \\
\hline
GPT-4o (w/o Graph)     & 61.54 (8)  & 46.15 (6)  & 30.77 (4) \\
GPT-4o (w/ Graph)      & \textbf{69.23 (9)}  & \textbf{61.54 (8)} & \textbf{46.15 (6)} \\
\hline
LLaMA 3.2 (w/o Graph)  & 46.15 (6)  & 23.08 (3)  & 15.38 (2) \\
LLaMA 3.2 (w/ Graph)   & \textbf{53.85 (7)}  & \textbf{38.46 (5)} & \textbf{30.77 (4)} \\
\hline
\end{tabular}
\end{table}

\section{Limitations and Future Work}

\label{sec:limitations}

While our system demonstrates robust performance on navigation tasks, several limitations indicate clear directions for future work. Our floorplans analysis framework, powered by a language model, generates node maps and adjacency graphs that facilitate effective navigation. However, achieving robust real-world deployment requires developing an accessible solution that functions effectively across various indoor environments. This is particularly important when compared to existing infrastructure-dependent solutions, such as BLE beacons or RFID systems. For practical deployment on mobile and wearable devices, we must consider the trade-off between accuracy and computational overhead. In this study, we focus on floorplans parsing and navigation tasks and do not address obstacle avoidance. Our proposed workflow generates human-like navigation that facilitates an efficient wayfinding for the user. However, it is essential to note that the lack of obstacle avoidance, particularly for dynamic obstacles, decreases the safe navigation of blind users. From a user perspective, we still need extensive evaluation across different architectural floorplans and BLV participants to validate how well our system generalizes and performs reliably in different settings. The main limitations of this study include the user study on BLV people, which can impact the model's performance, as we conducted this experiment with one sighted user and need to extend it to a group of BLV people to analyze the model's robustness in real-world environments. 

We plan to conduct IRB-approved research with BLV participants to evaluate system accessibility and user experience, which ensures our approach meets the real needs of the community we aim to serve.  Looking ahead, our future work focuses on several key areas. First, we are working to develop a more lightweight workflow that can be applied to integrate on mobile devices. Moreover, we are exploring adaptive prompt-tuning techniques to generate more human-like navigation instructions tailored to user preferences in unfamiliar environments. In the near future, we plan to evaluate Floorplan2Guide in more complex, larger indoor spaces to understand its scalability and performance across different floorplans.

\section{Conclusion}
\label{sec:conclusion}
We introduced Floorplan2Guide, an LLM-guided pipeline that assists BLV users to provide safe navigation in indoor environments by interpreting floorplans and generating context-aware navigation descriptions. Our workflow combines computer vision with foundation models to transform floorplans into a knowledge graph to provide step-by-step directions to the BLV users. Real-world tests at the UMBC Math \& Psychology building showed that our navigation system works well in practice, generating instructions that align with human-annotated routes. One key advantage of our approach is that it requires only the floorplans and a camera-equipped device to scan ArUco markers, instead of relying on infrastructure such as BLE beacons, RFID, or similar systems. Constructing a spatial knowledge graph can reduce hallucinations in topological reasoning and make path planning more reliable and precise for BLV users. Experimental results show that few-shot prompting improves navigation accuracy across different models and that language models can create accessible navigation systems that help BLV users navigate with confidence and safely. 

\section{Ethical Considerations}
LLMs were employed solely for language refinement; the authors conducted all experimental design and analysis.

\newpage
\appendix
\section{Appendix}
\label{append}

In this section, the visualization of the MP-1 floorplan adjacency graph and detailed system prompts with few-shot learning examples are presented for spatial knowledge graph generation and navigation instruction tasks.

\begin{figure}[ht]
    \centering
    \includegraphics[width=0.98\columnwidth,height=0.45\textheight,keepaspectratio]{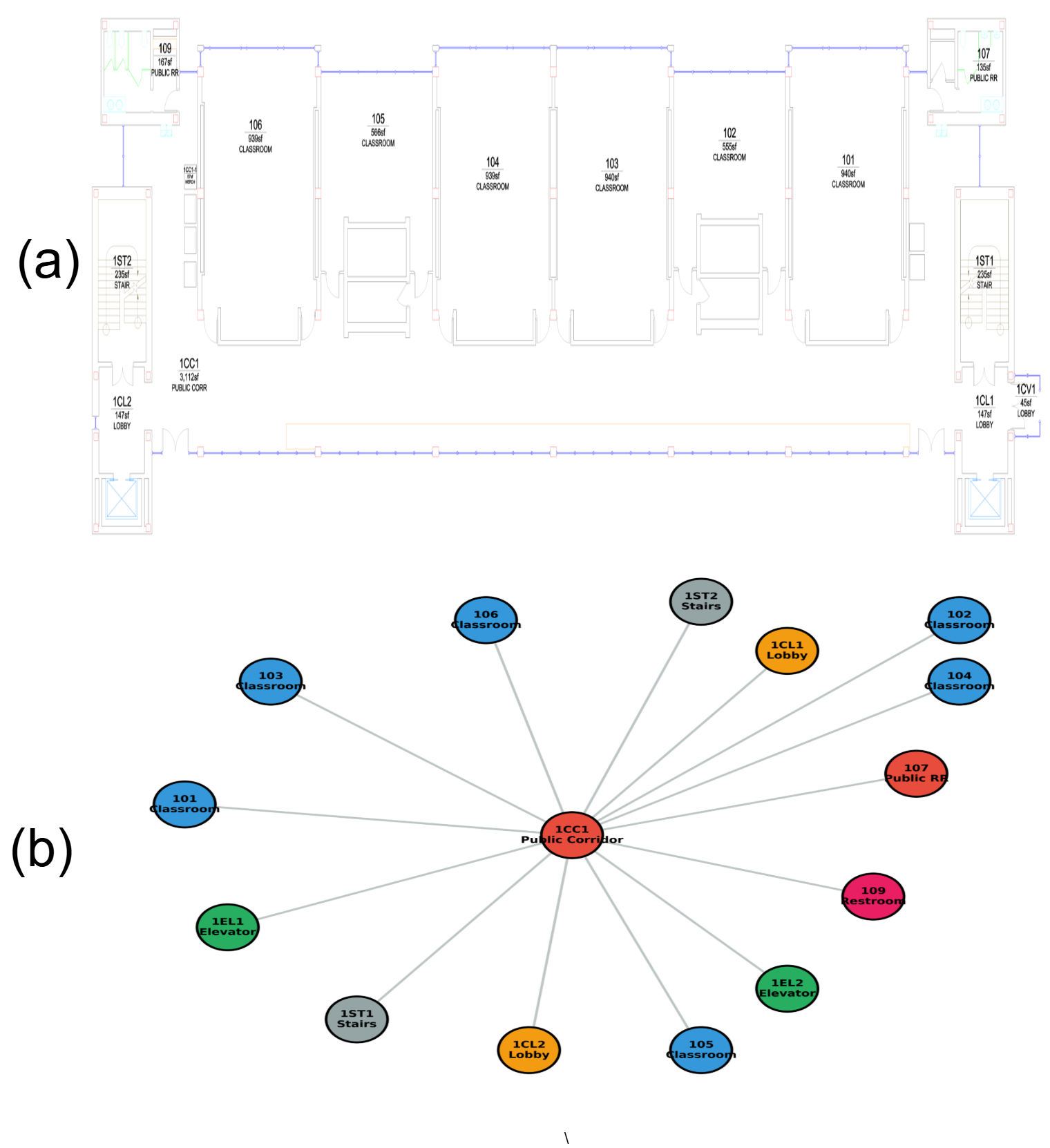}
    \caption{(a) Examples of floor plans from the MP-1 dataset, and (b) their corresponding adjacency graph representations.}
    \label{fig:spatial_graph_mp1}
\end{figure}

\subsection{Spatial Knowledge Graph Generation Prompt}

\begin{tcolorbox}[
    title=System Prompt,
    colback=gray!5,
    colframe=black!75,
    boxrule=1pt,
    arc=2pt,
    left=2mm,
    right=2mm,
    top=1.5mm,
    bottom=1.5mm,
    breakable
]
\small
You are a multimodal reasoning system that specializes in floorplan parsing. Your task is to analyze the provided floorplan image, along with OCR text and CAD metadata, to extract geometric, semantic, and spatial relationships between architectural components. Generate a spatial knowledge graph from a floorplan using reasoning over visual, OCR, and scale factor from floorplan/CAD metadata.

\vspace{1mm}
\noindent\textbf{Inputs:} Floorplan image, OCR text, and CAD metadata (scale factor).

\vspace{1mm}
\noindent\textbf{Reasoning Levels:}

\vspace{1mm}
\noindent\textbf{1. Perception: Identify Visual Elements}
\begin{itemize}[itemsep=0mm, leftmargin=3mm, topsep=0mm]
  \item Detect walls, doors, boundaries, and text regions.
  \item Extract textual labels and dimensions from OCR.
  \item Recognize room outlines, corridors, and major circulation zones.
\end{itemize}

\vspace{1mm}
\noindent\textbf{2. Entity Grounding – Define and Label Components}
\begin{itemize}[itemsep=0mm, leftmargin=3mm, topsep=0mm]
  \item Match OCR labels to regions (e.g., "Room 201" $\rightarrow$ region polygon).
  \item Classify entity types: room, hallway, entrance, stair, elevator.
  \item Assign unique IDs (\texttt{ROOM\_001}, \texttt{HALL\_001}, etc.).
  \item Compute geometric info: centroid, area, bounding box, coordinates (apply scale factor).
\end{itemize}

\vspace{1mm}
\noindent\textbf{3. Spatial Relations – Connect Entities}
\begin{itemize}[itemsep=0mm, leftmargin=3mm, topsep=0mm]
  \item For each visible door, find adjacent entities.
  \item Record connection pairs $(E_i, E_j)$ with basic properties: width, type, swing direction.
  \item Orient layout (North = top); define directional references.
\end{itemize}

\vspace{1mm}
\noindent\textbf{4. Graph Construction – Build Structural Representation}
\begin{itemize}[itemsep=0mm, leftmargin=3mm, topsep=0mm]
  \item Create a node list with all entities and attributes.
  \item Build an adjacency matrix or edge list for connections.
  \item Ensure each connection is bidirectional and unique.
  \item Integrate all information into a coherent graph format.
\end{itemize}

\vspace{2mm}
\noindent\textbf{Integration and Output:}
Combine perception, entities, and relations into JSON: \texttt{nodes[]} (entities with IDs, geometry, attributes), \texttt{edges[]} (connections), \texttt{adjacency\_matrix[][]} (binary connectivity), \texttt{metadata} (scale, units, orientation). Output must be valid JSON only.

\vspace{1mm}
\noindent\textbf{Core Constraints:}
OCR labels override visual inference. Only visible doors create edges. Node IDs are unique and consistent. Graph connections must be symmetric. Verify each edge corresponds to a valid node pair.

\end{tcolorbox}

\vspace{2mm}
\begin{tcolorbox}[
    title=Response Example,
    colback=gray!5,
    colframe=black!75,
    boxrule=0.8pt,
    arc=2pt,
    left=2mm,
    right=2mm,
    top=1.5mm,
    bottom=1.5mm,
    fontupper=\small,
    breakable
]

\begin{verbatim}
{
  "nodes": [{
    "id": "room_109",
    "type": "restroom",
    "label": "109",
    "coordinates": [113, 65],
    "coordinates_real_m": [1.13, 0.65],
    "geometry": {
      "area_m2": 8.0,
      "bbox": 
      [[73,35],[153,35],[153,95],[73,95]],
      "width_m": 0.8,
      "length_m": 0.6
    },
    "marker_id": "ROOM_109",
    "attributes": {
      "function": "restroom",
      "accessible": true,
      "gender": "unspecified"
    },
    "adjacent": ["room_1CC1"]
  }],
  "edges": [{
    "from": "room_109",
    "to": "room_1CC1",
    "via_door": "D013",
    "door_width_m": 0.9,
    "connection_type": "doorway"
  }],
  "adjacency_matrix": [[0, 1], [1, 0]],
  "metadata": {
    "scale_factor": 0.01,
    "units": "meters",
    "orientation": "north_up"
  }
}
\end{verbatim}

\end{tcolorbox}

\subsection{Navigation Instruction}

\begin{tcolorbox}[
    title= System Prompt,
    colback=gray!5,
    colframe=black!75,
    boxrule=1pt,
    arc=2pt,
    left=2mm,
    right=2mm,
    top=1.5mm,
    bottom=1.5mm,
    breakable
]
\small
You are an intelligent indoor navigation assistant designed to guide blind or low-vision (BLV) users through indoor environments using a spatial knowledge graph extracted from floorplans. Generate step-by-step navigation instructions using the spatial knowledge graph derived from floorplan analysis.

\vspace{1mm}
\noindent\textbf{Navigation Parameters:}
\begin{itemize}[itemsep=0mm, leftmargin=3mm, topsep=0mm]
  \item Step size: 40 cm (approximately one stride)
  \item Available actions: move forward, turn left, turn right, turn around, stop
  \item Perspective: First-person directional guidance
  \item Priority: Safety and sensory feedback
\end{itemize}

\vspace{1mm}
\noindent\textbf{Instruction Requirements:}
\begin{itemize}[itemsep=0mm, leftmargin=3mm, topsep=0mm]
  \item Provide specific step counts for each movement segment
  \item Include tactile and spatial landmarks for confirmation
  \item Describe environmental features at shoulder/hand height
  \item Ensure instructions comply with adjacency matrix constraints
  \item Highlight potential hazards and accessibility features
\end{itemize}

\vspace{2mm}
\noindent\textbf{Output Structure:}
The system generates structured navigation data containing: \texttt{navigation\_metadata} (start location, destination, total path length), \texttt{navigation\_instructions[]} (sequential steps with action, current position, confirmation landmarks, and \texttt{safety\_assessment} (hazard level, critical warnings, and accessibility features).

\vspace{1mm}
\noindent\textbf{Core Constraints:}
All instructions must be executable from a first-person perspective. Each step must include confirmation landmarks. Safety warnings must be comprehensive and prioritized. Path must respect the connectivity defined in the spatial knowledge graph.

\vspace{2mm}

\vspace{2mm}

\textit{Input : }   Sample Floorplan, Knowledge Graph, User Query, shot Examples

\begin{verbatim}

\end{verbatim}

\textit{User Query:} "I need to use the restroom, I am on CHAMBER1 on according to the floorplan"

\textit{Generated Navigation Instructions:}
\begin{verbatim}
{
  "navigation_metadata": {
    "start_location": "CHAMBRE1",
    "destination": "WC",
    "estimated_distance": "6.3 meters",
    "accessibility_level": "high"
  },
  "navigation_instructions": [
    {
      "step": 1,
      "action": "move forward 3 steps"
    
    },
    {
      "step": 2,
      "action": "exit through door D01"
    },
    {
      "step": 3,
      "action": "turn left 90 degrees",
  
      "current_position": "HALL corridor"
    },
    {
      "step": 4,
      "action": "move forward 10 steps"
    },
    {
      "step": 5,
      "action": "turn right 90 degrees",
      "current_position":"facing WC entrance"
    },
    {
      "step": 6,
      "action": "move forward 5 steps"
    },
    {
      "step": 7,
      "action": "stop - destination reached",
      "current_position": "at WC entrance"
    }
  ],
  "safety_assessment": {
    "hazard_level": "low",
    "critical_warnings": [
      "Door D01 may swing inward"
    ],
    "accessibility_features": [
      "Wide hallway",
      "Tactile WC signage available"
    ]
  }
}
\end{verbatim}

\vspace{2mm}
\noindent\textit{\textbf{Note:}} For 3-shot prompting, two more examples need to be added to cover different route types and complex routes with multiple decision points. For 5-shot prompting, four additional examples should be added to cover diverse architectural layouts and route types.

\vspace{2mm}

\vspace{2mm}
\noindent\textbf{Target Query:}

\textit{Input Floorplan:} [Target floorplan image]

\textit{ Knowledge Graph:} [Target graph structure]

\textit{User Query:} [User's navigation request]

\textit{Task:} Generate complete navigation instructions following the demonstrated format.

\end{tcolorbox}

\end{document}